\documentclass[a4paper, 10 pt, conference]{ieeeconf}  

\IEEEoverridecommandlockouts                              

\usepackage{graphicx}
\usepackage{multirow}
\usepackage{amsmath,amssymb,amsfonts}
\usepackage{mathtools}
\usepackage{adjustbox}
\usepackage{comment}

\title{\LARGE \bf
Improving Graph Machine Learning Performance Through Feature Augmentation Based on Network Control Theory
}

\author{Anwar~Said$^{1}$, Obaid~Ullah~Ahmad$^{2}$, Waseem~Abbas$^{3}$, Mudassir~Shabbir$^{4}$,  Xenofon~Koutsoukos$^{1}$ 
\thanks{*This material is based upon work supported by the National Science Foundation Grant Nos. 2325416, and 2325417.}
\thanks{$^{1}$Anwar Said, and Xenofon Koutsoukos are with the Computer Science Department at the Vanderbilt University, Nashville, TN. Email: 
{\tt\small anwar.said@vanderbilt.edu, xenofon.koutsoukos@vanderbilt.edu}}
\thanks{$^{2}$Obaid Ullah Ahmad is with the Electrical Engineering Department at the University of Texas at Dallas, Richardson, TX 75080. Email: {\tt\small Obaidullah.Ahmad@utdallas.edu}} 
\thanks{$^{3}$Waseem Abbas is with the Systems Engineering Department at the University of Texas at Dallas, Richardson, TX. Email: {\tt\small waseem.abbas@utdallas.edu}}
\thanks{$^{4}$Mudassir~Shabbir is with the Computer Science Department at Information Technology University, Lahore, Pakistan Email: {\tt\small mudassir.shabbir@itu.edu.pk}}
\thanks{The 31st Mediterranean Conference on Control and Automation, 2024}
}

\begin{document}

\maketitle
\thispagestyle{empty}
\pagestyle{empty}

\begin{abstract}

Network control theory (NCT) offers a robust analytical framework for understanding the influence of network topology on dynamic behaviors, enabling researchers to decipher how certain patterns of external control measures can steer system dynamics towards desired states. Distinguished from other structure-function methodologies, NCT's predictive capabilities can be coupled with deploying Graph Neural Networks (GNNs), which have demonstrated exceptional utility in various network-based learning tasks. 
However, the performance of GNNs heavily relies on the expressiveness of node features, and the lack of node features can greatly degrade their performance. Furthermore, many real-world systems may lack node-level information, posing a challenge for GNNs.
To tackle this challenge, we introduce a novel approach, NCT-based Enhanced Feature Augmentation (NCT-EFA), that assimilates average controllability, along with other centrality indices, into the feature augmentation pipeline to enhance GNNs performance. Our evaluation of NCT-EFA, on six benchmark GNN models across two experimental setting—solely employing average controllability and in combination with additional centrality metrics—showcases an improved performance reaching as high as 11\%. Our results demonstrate that incorporating NCT into feature enrichment can substantively extend the applicability and heighten the performance of GNNs in scenarios where node-level information is unavailable.

\end{abstract}

\section{INTRODUCTION}
\label{sec:introduction}

Network Control theory (NCT), with its rigorous mathematical foundation and practical applicability, has been instrumental in crafting systems that respond predictably to inputs, ensuring stability, efficiency, and desired performance \cite{yaziciouglu2022strong}. Its principles permeate modern engineering, deftly guiding everything from simple home appliances to sophisticated aerospace vehicles \cite{friedland2012control}. 
Enter graph machine learning, an emerging field that thrives on the abstraction and analysis of relational data, inherently capturing the intricacies of complex systems as networks \cite{hamilton2020graph}. By intertwining the predictive prowess of graph-based models with the robust framework of control theory, there lies a profound opportunity to revolutionize how we understand, design, and manage dynamic systems \cite{said2023network}.

Average controllability is a critical metric in systems theory that quantifies the ability of a node within a network to influence the system's overall behavior through the introduction of external inputs \cite{kalman1960contributions}. It captures the essence of a node's role in controlling the dynamical landscape of a network by measuring the system's response to impulses, effectively gauging a node's intrinsic capability to steer the system's states. 

As such, this metric is invaluable for discerning the underlying structural properties that dictate how control energy proliferates through the network, offering a window into the architecture's innate information flow.

Building on the insights provided by average controllability within the purview of NCT, network science on the other hand offers an expansive framework to further dissect and enhance our understanding of systemic influence and regulation \cite{barabasi2013network}. Centrality measures within network science have emerged as pivotal tools in this venture, quantifying the roles and significance of individual nodes amid the complex web of interactions that typify controlled systems. Closeness centrality reveals those nodes that efficiently affect the entire network, due to their minimized path lengths to all other nodes, underpinning their strategic value for swift and comprehensive system-wide influence \cite{said2018cc}. Eigenvector centrality takes into account the notion that connections to highly influential nodes contribute more significantly to a node's importance, emphasizing the network's inherent hierarchy and influence distribution. Meanwhile, betweenness centrality captures the nodes that frequently serve as bridges in the shortest paths between others, highlighting their critical role in facilitating or bottling the flow of control through the network \cite{freeman2002centrality}. These centrality metrics, thus, encapsulate vital information about the network structure, and play vital role in many modeling tasks \cite{parkes2021network}. 

Graph Neural Networks (GNNs) have emerged as a sophisticated approach for performing machine learning on graph-structured data, harnessing the framework of message passing to learn refined node representations that are critical for a variety of downstream machine learning tasks \cite{kipf2016semi}. The message passing mechanism - central to the functionality of GNNs - iteratively updates node representations by aggregating features from their respective neighborhoods, thereby encapsulating both local and global structural information within the graph \cite{hamilton2017inductive}. Nevertheless, the performance of GNNs is intrinsically linked to the initial availability and quality of the node features; deficient feature sets can significantly curtail the representational capacity of the learned embeddings \cite{said2023enhanced}. A prevalent challenge arises in circumstances where node features are sparse or entirely missing. Under such constraints, the customary practice involves the utilization of one-hot encoding of node degrees as surrogate features, injecting a rudimentary form of structural information into the model \cite{xu2018powerful}. Despite this adaptation, the absence of rich, discriminative features often results in GNNs that struggle to achieve the depth of understanding necessary for complex inference tasks, thereby revealing a limitation in their ability to thoroughly exploit the intricate relational patterns present within graph data \cite{gilmer2017neural}.

To address the challenge of feature impoverishment in GNNs, our study proposes leveraging network controllability and centrality measures to enrich node feature sets, potentially improving GNN performance. Our main contributions are as follows:
\begin{itemize}

\item \textbf{Feature augmentation:} We propose integrating metrics like average controllability, betweenness centrality, closeness centrality, and eigenvector centrality as node features within the GNN framework.

\item \textbf{Unique representation of average controllability:} We use a new node representation method to encode average controllability as a node feature. Through an ablation study, we demonstrate the effectiveness of the proposed scheme in enhancing classification performance.

\item \textbf{Evaluation and comparison:} In social network classification domain, where node features are missing, we evaluate several GNN models using our proposed approach. We compare these models with baseline techniques such as one-hot-degree encoding. Our results demonstrate the superior performance of our proposed method.
\end{itemize}
By incorporating these additional metrics and exploring novel representations, our study aims to enhance the capabilities of GNNs and address the challenge of missing feature in network analysis tasks.

\section{Related Work}
\label{sec:related-work}

The field of integrating NCT with graph machine learning is still in its early stages, with few studies exploring this intersection. To the best of our knowledge, only one study has been proposed that uses NCT metrics to derive graph representations for downstream graph classification tasks \cite{said2023network}. However, no existing study has integrated NCT with GNNs to advance the field. In the following, we briefly present \cite{said2023network} and provide references for further reading to interested readers. Due to space limitations, we keep the related work section concise.

The authors in \cite{said2023network} introduce an intriguing approach to integrate NCT with graph machine learning. The core idea of the paper involves using various controllability metrics to derive expressive graph representations. Specifically, this study explores the controllability gramian as a source of controllability information. It considers metrics such as trace, rank, and the first and last three eigenvalues of the controllability gramian as potential features for graph representations. Moreover, it uses different numbers of leaders and repeats the process to derive an expressive set of features for the final graph representation. This study serves as a catalyst for further research to explore the potential of integrating NCT with GNNs, opening new avenues for advancing the field of graph machine learning.

In the realm of graph machine learning, the field is characterized by a rapid pace of research, primarily focusing on two main approaches: graph embedding methods and GNNs. Graph embedding methods employ various graph-theoretic and statistical techniques to derive representations of graphs, which are then utilized independently to train machine learning models such as support vector machines and random forests \cite{kriege2020survey,said2023survey}. A few notable and recent graph embedding methods include \cite{togninalli2019wasserstein,kondor2016multiscale,said2023augmenting}. On the other hand, GNNs are deep learning-based approaches that are trained in an end-to-end fashion. They utilize techniques such as message passing, spectral methods, and recurrent neural networks to learn representations at the node and graph levels. GNNs offer an advantage over embedding methods by effectively leveraging both the graph's topology and its node and edge features \cite{bresson2017residual,xu2018powerful}. For further exploration of these concepts, interested readers are encouraged to refer to \cite{hamilton2020graph,kipf2016semi,velivckovic2017graph,wu2020comprehensive}.

\section{Methodology}
\label{sec:method}

NCT is a powerful paradigm originally rooted in the domain of systems and control engineering, which has found its footing in the study of social networks by offering a sophisticated mathematical framework for understanding how influence and information propagate through complex relational systems. When applied to social networks, NCT can provide a nuanced perspective on the capacity of individual nodes, or users 
within the network, to affect overall network behavior, based on the topology and dynamics of social interactions \cite{parkes2023using}.

\subsection{Network Controllability Metrics}
\label{sec:nct-metrics}

By identifying key control nodes within a social network—akin to the influential nodes in a social network—NCT enables us to tailor node representations such that they encapsulate not only the structural attributes of these entities but also their potential to exert control over the network state. This could include their capability to disseminate information effectively or to reconfigure connections for desired social outcomes strategically. Employing NCT to parameterize and analyze control nodes allows for the enhancement of node features, which can, in turn, improve the performance of models on downstream tasks such as social influence prediction, node classification, and graph classification, among others.

The structural composition of NCT encapsulates a network defined by the adjacency matrix $A \in \mathbb{R}^{N \times N}$ of $N = |V|$ nodes and a control set $B \in \mathbb{R}^{N \times m}$. Within this framework, it posits that the temporal evolution of the state of any given node, denoted $x_i(t)$, is governed by a composite function. This function represents the cumulative influence of all preceding nodes, expressed as $x_j(t)$, in conjunction with any external inputs, $u(t)$, each modulated by appropriate weights within the network's topology. When the progression of node states is conceptualized through the lens of rates of change, such that the activity from preceding nodes influences the ongoing rate of state alteration in subsequent nodes, the model adopts the construct of a differential equation presented as follows:

\begin{equation}
    \label{eq:control}
    \frac{d}{dt}\boldsymbol{x}(t) = A\boldsymbol{x}+B\boldsymbol{u}(t)
\end{equation}

where $\boldsymbol{x}(t) = [x_1(t), x_2(t),\ldots, x_n(t)]^\top$ is the vector of node states, $A$ is the adjacency matrix, and $\boldsymbol{u}(t) = [u_1(t), u_2(t),\ldots, u_n(t)]^\top$ is the vector of control signals. $B^{N \times m}$ quantifies the effects of inputs on each node. In our experiments, the control set matrix $B^{N \times N}$ is defined as the identity matrix, representing a uniform full control set.

The controllability Gramian is a powerful mathematical concept that plays a crucial role in understanding the control behavior of a network \cite{pasqualetti2014controllability,ahmad2024control}. By utilizing the controllability Gramian, we can measure the ease with which we can transition from one state to another in terms of the necessary control energy. In the context of the system defined in Equation \ref{eq:control}, the infinite horizon controllability Gramian can be formally expressed as follows:

\begin{figure*}[!t]
    \centering
    \includegraphics[width=\textwidth]{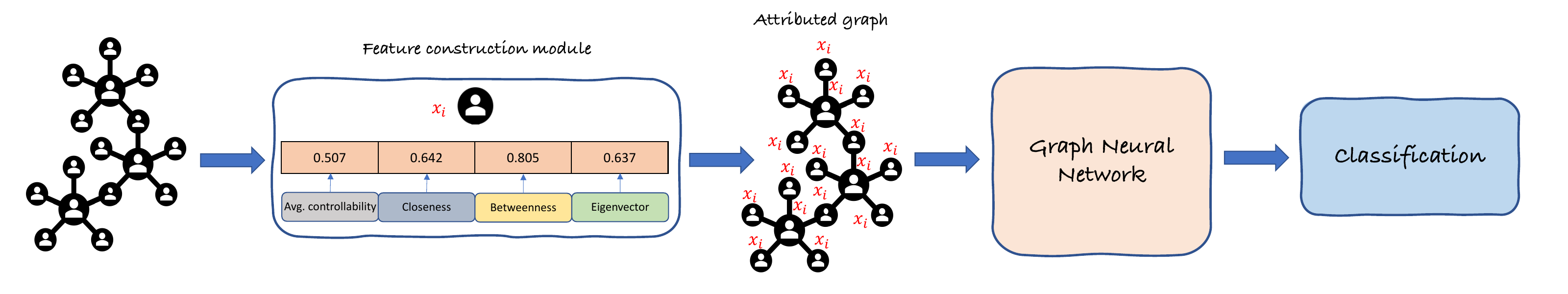}
    \caption{Illustration of the proposed NCT-EFA framework. Starting with a simple graph, NCT-EFA utilizes network control theory and network science metrics to enhance the graph by node features. These graphs with enriched features are subsequently fed into GNN module for training and the downstream classification task.}
    \label{fig:architecture}
\end{figure*}

\begin{equation}
    \label{eq:Gramian}
    \mathcal{W} = \int_{0}^{\infty}e^{-A\tau}(-B)(-B)^\top e^{-A^\top \tau}d\tau \; \in \; \mathbb{R}^{N_f \times N_f}.
\end{equation}

When the system is stable, implying that all eigenvalues of $-A$ possess negative real parts, $\mathcal{W}$ converges asymptotically and can be determined through the Lyapunov equation.

\begin{equation}
    \label{eq:Lyapunov}
    (-A)\mathcal{W} + \mathcal{W}(-A)^\top + (-B)(-B)^\top  = 0,
\end{equation}

The formulation presented in Equation \ref{eq:control} provides us with a framework to derive various NCT metrics from the observed system—among these, a key parameter of interest is average controllability, which is the focal point of the current study.

\textbf{Average Controllability:} Average controllability quantifies the extent to which the state vector, $\boldsymbol{x}(t)$, can be modulated in response to input stimuli administered to an individual node within the network~\cite{parkes2021network}. A node that exhibits a higher degree of average controllability possesses an enhanced capacity to utilize the underlying graph structure to disseminate an impetus across the entire network. Average controllability is defined as the trace of the controllability Gramian, expressed as:
\begin{equation*}
\label{eq:avg-cont}
        \mathcal{C}_a = tr(\mathcal{W}) 
\end{equation*}

Next, we detail GNNs, that will be used as a learning architecture for social network classification task.  


\subsection{Graph Neural Networks}
\label{subsec:gnns}

GNNs represent an innovative class of neural networks specifically designed to operate on graph-structured data \cite{hamilton2020graph}. GNNs are particularly adept at learning from graph-based data due to their ability to capture the inherent relationships and interdependencies between nodes through a process of iterative information aggregation \cite{kipf2016semi}. Unlike traditional neural networks that assume independent and identical distribution of data, GNNs embrace the irregularities and topological characteristics of graphs, enabling them to learn and make predictions on data that exhibits rich relational context. This capability makes GNNs the preferred choice for a multitude of applications ranging from social network analysis, brain networks, and recommendation systems to traffic network flow optimization, where relational patterns are of paramount importance \cite{said2023neurograph,hamilton2020graph}.

Building upon the foundation of GNNs, Message Passing Neural Networks (MPNNs) extend this concept with a specific focus on node features and their interactions. MPNNs operate on the principle that each node's feature representation is updated by recursively aggregating features from its neighbors, thereby encapsulating both local structures and global context within the graph. This process is formalized by the message passing equation:
\begin{equation*}
\small
    \boldsymbol{h}_v^{(t+1)} = \text{UPDATE} \left( \boldsymbol{h}_v^{(t)}, 
    \text{AGGREGATE} \left( \boldsymbol{h}_u^{(t)} : u \in \mathcal{N}(v) \right) \right) 
\end{equation*}
Here, $\boldsymbol{h}_v^{(t+1)}$ represents the updated feature vector of node $v$ at iteration $(t+1)$; $ \text{UPDATE} $ and $ \text{AGGREGATE} $ are functions that define the updating and aggregation mechanisms, respectively; $\boldsymbol{h}_u^{(t)} $ is the feature vector of a neighboring node $u$; and $ \mathcal{N}(v) $ denotes the set of neighbors of node $v$. The power of MPNNs lies in the expressiveness of learned node representations, which can significantly enhance model performance on downstream tasks. Consequently, integrating better features through improved UPDATE and AGGREGATE functions can result in more expressive GNN models, capable of capturing complex patterns and ultimately leading to more accurate predictions.
\vspace{-0.05in}
\subsection{Constructing Node Features}
\label{subsec:node-features-construction}

In many real-world networks, the absence of node features is commonplace, attributable to a variety of factors including privacy constraints, data collection errors, proprietary information protection, and instances where features may be unobserved or inherently non-existent. In situations where node attributes are lacking, strategies like one-hot-degree encoding or the employment of randomly assigned features are often utilized to facilitate the training of MPNNs. However, the performance of these models can be compromised due to the paucity of informative features, a limitation corroborated by numerous preceding studies \cite{said2023enhanced}. To mitigate this challenge and endow graphs devoid of node features with enriched representations, we advocate for the integration of NCT which has been described in section \ref{sec:nct-metrics}. In the following section, we detail the rest of the network science measures that we use for enriching the node features.


\textbf{Closeness Centrality:} Closeness centrality is a measure used in network analysis to identify the relative importance of a node within a graph, based on its proximity to all other nodes. It is calculated as the reciprocal of the sum of the shortest path distances from a given node to all other nodes in the network. Nodes with higher closeness centrality are seen as having better potential to quickly interact with all others due to their central positioning in the graph's structure.

\textbf{Betweenness Centrality:} Betweenness centrality quantifies the significance of a node within a network by measuring the extent to which the node acts as a bridge along the shortest paths between other nodes. It is computed as the fraction of all-pairs shortest paths that pass through a given node, indicating its role as an intermediary in facilitating communication or connectivity within the network. Nodes with high betweenness centrality are often crucial for the flow of information, resources, or influence across the network, as they can control and influence the interactions between different parts of the graph.

\textbf{Eigenvector Centrality:} This is a measure of influence within a network that assigns relative scores to all nodes based on the principle that connections to high-scoring nodes contribute more to the score of a node than equal connections to low-scoring nodes. This centrality considers not only the number of connections or edges a node has but also the quality of those connections in terms of the centrality of its neighbors. It is calculated by finding the eigenvector corresponding to the largest eigenvalue of the network's adjacency matrix, with the components of this principal eigenvector giving the centrality scores of the nodes.

Building upon the four central measures—we compute these metrics for each individual node within the network and collate them into a feature vector representative of the node's structural characteristics. This vector of network attributes effectively captures the node's connectivity profile, potential for information dissemination, and overall influence within the network structure. To aid in understanding and visualizing this methodology, an illustrative representation of how these features are computed and assembled for every node has been shown in Figure \ref{fig:architecture}.

\textbf{Ranks one-hot encoding:}
In various systems, there's often a need to introduce external inputs or amplify the influence of key users to alter the system's behavior. One example could be considering the influential users in a social network.  In such cases, traditional network science measures that focus solely on graph topology may not suffice. Instead, we may rely on average controllability as a node feature. However, using a single real number as a feature may not be optimal for the model's performance. To address this, we propose encoding average controllability in the following manner.

Given the average controllability vector for the entire graph, computed as described in \cite{parkes2023using} where each value corresponds to a node in the graph, we create a histogram $\mathcal{H}$ with $k$ bins to represent the distribution. Each bin corresponds to a range of average controllability values, and the height of each bin represents the frequency of nodes with average controllability values falling within that range. This histogram provides a summary of the controllability distribution across nodes in the graph, highlighting the prevalence of certain controllability levels. To create a feature vector for a node $v$ using $\mathcal{H}$, we one-hot encode the feature vector based on the index $\mathcal{H}(i)$ where the average controllability of node $v$ falls. Formally, the one-hot encoding for the feature vector $\mathbf{h}^0_v$ can be expressed as:

\begin{equation*}
\small
\mathbf{h}^0_v(i) = 
\begin{cases} 
1 & \text{if} \hspace{0.05in} \mathcal{C}_a(v) \in \mathcal{H}(i) \\
0 & \text{otherwise}
\end{cases}
\end{equation*}

The final component of our framework is dedicated to learning over the attributed graph. Within this context, we may employ any message passing-based GNN variant and adhere to conventional training protocols to fine-tune the model. In training our model for the graph classification task, we employ a binary cross-entropy loss to optimize the learning process. After training, the model is then applied to the binary classification task, using its learned features to distinguish between two separate classes.

\section{Numerical Evaluation}
\label{sec:evaluation}



\begin{table*}[!t]
\centering
\caption{Dataset stats}
\begin{tabular}{|l|c|cc|cc|cc|c|c|}
\hline
\multirow{2}{*}{\textbf{Dataset}} & \multirow{2}{*}{\textbf{Graphs}} & \multicolumn{2}{c|}{\textbf{Nodes}}     & \multicolumn{2}{c|}{\textbf{Density}}       & \multicolumn{2}{c|}{\textbf{Diameter}} &\textbf{Classes} & \textbf{Task}\\ \cline{3-8} 
                         &                         & \multicolumn{1}{c|}{Min} & Max & \multicolumn{1}{c|}{Min}   & Max   & \multicolumn{1}{c|}{Min} & Max &&\\ \hline
Reddit Threads           & 203,088                 & \multicolumn{1}{c|}{11}  & 97  & \multicolumn{1}{c|}{0.021} & 0.382 & \multicolumn{1}{c|}{2}   & 27  &2&Graph Classification\\ \hline
GitHub Stargazers        & 12,725                  & \multicolumn{1}{c|}{10}  & 957 & \multicolumn{1}{c|}{0.003} & 0.561 & \multicolumn{1}{c|}{2}   & 18  &2&Graph Classification\\ \hline
\end{tabular}
\label{tab:dataset-stats}
\end{table*}

\begin{table*}[!t]
\tiny
\centering
\caption{Comparison of ROC AUC scores of the proposed method against one-hot-degree encoding (deg)}
\begin{tabular}{|l|cc|cc|cc|cc|cc|cc|}
\hline
\multirow{2}{*}{Datasets} & \multicolumn{2}{c|}{\textbf{k-GNN}}      & \multicolumn{2}{c|}{\textbf{SAGE}}        & \multicolumn{2}{c|}{\textbf{GCN}}   & \multicolumn{2}{c|}{\textbf{UniMP}}   & \multicolumn{2}{c|}{\textbf{ResGatedGCN}}   & \multicolumn{2}{c|}{\textbf{GAT}}     \\ \cline{2-13} 
                          & \multicolumn{1}{c|}{\textit{deg}} & NCT-EFA & \multicolumn{1}{c|}{\textit{deg}} & NCT-EFA & \multicolumn{1}{c|}{\textit{deg}} & NCT-EFA & \multicolumn{1}{c|}{\textit{deg}} & NCT-EFA & \multicolumn{1}{c|}{\textit{deg}} & NCT-EFA & \multicolumn{1}{c|}{\textit{deg}} & NCT-EFA   \\ \hline
Reddit Threads            & \multicolumn{1}{c|}{$83.72$}    &   $\boldsymbol{84.06}$   & \multicolumn{1}{c|}{$83.64$}    &     $\boldsymbol{83.90}$ &   \multicolumn{1}{c|}{$82.83$}    &  $\boldsymbol{83.79}$    &\multicolumn{1}{c|}{$83.83$}    &$\boldsymbol{83.96}$     & \multicolumn{1}{c|}{$84.01$}    &  $\boldsymbol{84.12}$  &   \multicolumn{1}{c|}{$\boldsymbol{83.88}$}    &   $83.86$     \\ \hline

GitHub Stargazers         & \multicolumn{1}{c|}{$71.21$}    & $\boldsymbol{79.03}$     & \multicolumn{1}{c|}{$66.27$}    &   $\boldsymbol{75.82}$   & \multicolumn{1}{c|}{$68.34$}    &   $\boldsymbol{74.51}$   & \multicolumn{1}{c|}{$68.39$}    &   $\boldsymbol{77.48}$   & \multicolumn{1}{c|}{$74.97$}    &  $\boldsymbol{79.37}$    & \multicolumn{1}{c|}{$64.21$}    &   $\boldsymbol{75.90}$    \\ \hline
\end{tabular}
\label{tab:results_all_features}
\end{table*}

In this section, we evaluate the efficacy of our proposed methodology under two distinct experimental settings. Initially, we undertake a comparative analysis between the comprehensive feature set—encompassing average controllability, closeness centrality, betweenness centrality, and eigenvector centrality—and the baseline one-hot-degree encoding method. This comparison aims to establish the superiority of our enriched feature set in capturing the intricate topological nuances of the graph data. Subsequently, we show the effectiveness of employing one-hot-encoded ranks derived from average controllability measures. This second setting allows us to assess the impact of prioritizing nodes based on their average controllability and to ascertain the resultant performance gains in our graph neural network model.

\subsection{Datasets}

We consider the following two social network datasets in our experimental setup.

\textbf{Reddit Threads:} This dataset comprises an assortment of threads extracted from the Reddit platform, all of which were gathered during the month of May 2018. It encompasses both discussion and non-discussion based threads, presenting a diverse range of community interactions. The task is to distinguish between threads that facilitate discussion and those that do not, thereby classifying the content based on its conversational nature and potential for user engagement \cite{karateclub}.

\textbf{GitHub Stargazers:} This is a social network dataset of developers who have starred notable machine learning and web development repositories on GitHub. The task involves classifying these networks to ascertain if they correspond to web development or machine learning repositories based on their stargazing activities \cite{karateclub}. 

Both of these datasets correspond to a graph classification task. We present the statistics of these datasets in Table \ref{tab:dataset-stats}.

\subsection{Experimental Setup and Results}

We consider six well-known baselines graph convolution methods that include $k-$GNN \cite{morris2019weisfeiler}, GraphSAGE \cite{hamilton2017inductive}, GCN \cite{kipf2016semi}, Transformer Convolution (UniMP) \cite{shi2020masked}, Residual Gated Graph ConvNets (ResGatedGCN) \cite{bresson2017residual} and Graph Attention Network (GAT) \cite{velivckovic2017graph}.

The proposed learning framework comprises three layers of GNNs, each containing $64$ hidden units. Post these layers, Sort Aggregation \cite{zhang2018end} is applied. Following the aggregation step, we employ two layers of 1D convolution complemented by Max Pooling. This is succeeded by a multi-layer perceptron with two layers, each having 32 hidden neurons. For model evaluation, we resort to 10-fold cross-validation, training each model for a duration of 100 epochs. We set the learning rate at $1e^{-4}$ and the weight decay at $5e^{-2}$. All experiments are conducted on a Lambda machine equipped with an AMD Ryzen Threadripper PRO $5995WX 64-$Core CPU, 512 GB RAM, and an NVIDIA RTX 6000 GPU with 48 GB of memory.

\subsection{Results with Full Set of Features}

We present the ROC AUC (Receiver Operating Characteristic Area Under the Curve) classification results for both datasets using the full set of features in Table \ref{tab:results_all_features}. We maintained consistent architecture and experimental settings for evaluation, comparing the performance of both one-hot-degree encoding (deg) and the proposed method (NCT-EFA). Overall, we observed improved performance with NCT-EFA on both datasets, except for the GAT results on the Reddit dataset, which exhibited a slight $(0.02\%)$ decrease. Notably, NCT-EFA yielded an $11.69\%$ improvement with GAT on the GitHub Stargazers dataset, a $9.55\%$ improvement with GraphSAGE, and a $9.09\%$ improvement with UniMP. These results underscore the effectiveness of the proposed NCT-EFA approach for downstream classification tasks. It is worth noting that the relatively minor difference between one-hot-degree encoding and NCT-EFA on Reddit Threads dataset can be attributed to the dataset's graph topology playing a more significant role in classification than the node features.

\begin{figure}
    \centering
    \caption{Comparison of ROC AUC scores of the encoded ranks (average controllability) against degree one-hot-encoding. }
    \includegraphics[width=0.49\textwidth]{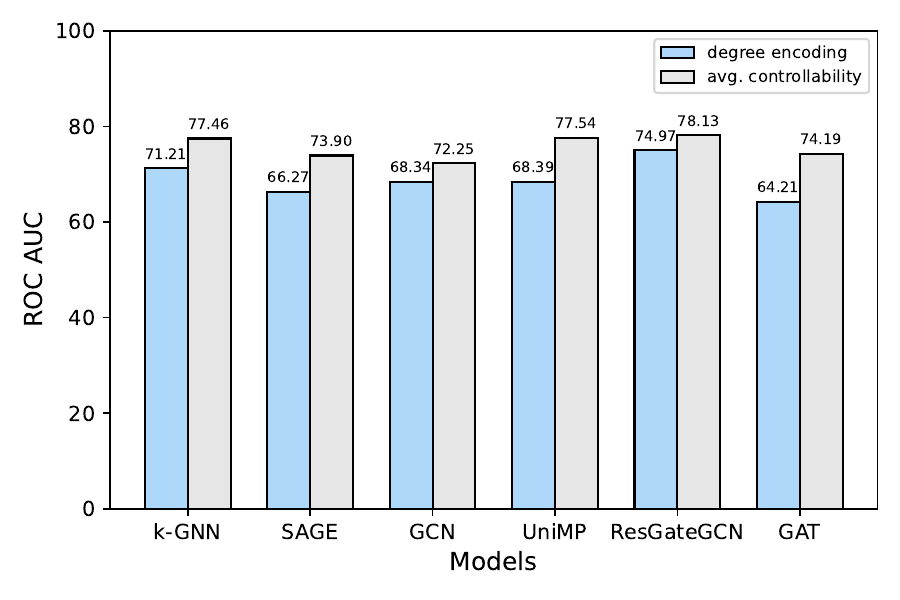}
    \label{fig:stargazers_results}
    \vspace{-0.5in}
\end{figure}

\subsection{Numerical Results with Average Controllability}

To illustrate the effectiveness of using only the average controllability, we conducted additional experiments. In this setup, we maintained the same architecture and experimental configuration but utilized one-hot encoded average controllability values, as discussed in Section \ref{subsec:node-features-construction}. We computed the average controllability for each node and then encoded these values to create the corresponding node features. Subsequently, we trained the same models using this encoded dataset and compared the results with those using one-hot-degree encoding. The results are presented in Figure \ref{fig:stargazers_results}.

Our results (ROC AUC) indicate that encoding average controllability significantly enhances the performance of all the GNN models. Specifically, we observed a $9.98\%$ improvement with GAT, $9.15\%$ with UniMP, and $7.63\%$ with GraphSAGE. Notably, the improvements seen with the full set of features in the previous section were $11.69\%$, $9.55\%$, and $9.05\%$, respectively. This suggests that average controllability contributes significantly more compared to the other features in these datasets.

\section{Conclusions and Future Work}
\label{sed:conclusion}

In this work, we have introduced a novel feature augmentation approach based on NCT, aimed at enhancing node features to boost the performance of GNNs in graph classification task within social networks. Our method leverages average controllability, closeness centrality, betweenness centrality, and eigenvector centrality as node features, which we evaluated with six different GNN architectures. Additionally, we introduced a novel rank encoding scheme that constructs expressive node features from average controllability, further enhancing the approach's effectiveness.

Our experimental results across various GNNs and a diverse social network dataset demonstrate consistent performance improvements, underscoring the efficacy of our proposed approach. These promising results motivate several key avenues for future exploration. For instance, combining various controllability metrics to augment GNNs could enhance their capabilities further. Additionally, integrating controllability metrics directly into the message passing mechanism of GNNs presents an intriguing research direction. Lastly, leveraging controllability metrics for constructing expressive graph embeddings represents another exciting path for future investigation.

The field of integrating controllability metrics with GNNs is relatively young but shows immense potential for developing more expressive graph machine learning methods. We believe that our work lays a foundation for exploring the integration of these two fields, paving the way for groundbreaking advancements in graph machine learning.

\bibliographystyle{IEEEtran}
\bibliography{references}

\begin{thebibliography}{10}
\providecommand{\url}[1]{#1}
\csname url@samestyle\endcsname
\providecommand{\newblock}{\relax}
\providecommand{\bibinfo}[2]{#2}
\providecommand{\BIBentrySTDinterwordspacing}{\spaceskip=0pt\relax}
\providecommand{\BIBentryALTinterwordstretchfactor}{4}
\providecommand{\BIBentryALTinterwordspacing}{\spaceskip=\fontdimen2\font plus
\BIBentryALTinterwordstretchfactor\fontdimen3\font minus \fontdimen4\font\relax}
\providecommand{\BIBforeignlanguage}[2]{{%
\expandafter\ifx\csname l@#1\endcsname\relax
\typeout{** WARNING: IEEEtran.bst: No hyphenation pattern has been}%
\typeout{** loaded for the language `#1'. Using the pattern for}%
\typeout{** the default language instead.}%
\else
\language=\csname l@#1\endcsname
\fi
#2}}
\providecommand{\BIBdecl}{\relax}
\BIBdecl

\bibitem{yaziciouglu2022strong}
Y.~Yaz{\i}c{\i}o{\u{g}}lu, M.~Shabbir, W.~Abbas, and X.~Koutsoukos, ``Strong structural controllability of networks: Comparison of bounds using distances and zero forcing,'' \emph{Automatica}, vol. 146, p. 110562, 2022.

\bibitem{friedland2012control}
B.~Friedland, \emph{Control system design: an introduction to state-space methods}.\hskip 1em plus 0.5em minus 0.4em\relax Courier Corporation, 2012.

\bibitem{hamilton2020graph}
W.~L. Hamilton, \emph{Graph representation learning}.\hskip 1em plus 0.5em minus 0.4em\relax Morgan \& Claypool Publishers, 2020.

\bibitem{said2023network}
A.~Said, O.~U. Ahmad, W.~Abbas, M.~Shabbir, and X.~Koutsoukos, ``Network controllability perspectives on graph representation,'' \emph{IEEE Transactions on Knowledge and Data Engineering}, 2023.

\bibitem{kalman1960contributions}
R.~E. Kalman \emph{et~al.}, ``Contributions to the theory of optimal control,'' \emph{Bol. soc. mat. mexicana}, vol.~5, no.~2, pp. 102--119, 1960.

\bibitem{barabasi2013network}
A.-L. Barab{\'a}si, ``Network science,'' \emph{Philosophical Transactions of the Royal Society A: Mathematical, Physical and Engineering Sciences}, vol. 371, no. 1987, p. 20120375, 2013.

\bibitem{said2018cc}
A.~Said, R.~A. Abbasi, O.~Maqbool, A.~Daud, and N.~R. Aljohani, ``Cc-ga: A clustering coefficient based genetic algorithm for detecting communities in social networks,'' \emph{Applied Soft Computing}, 2018.

\bibitem{freeman2002centrality}
L.~C. Freeman \emph{et~al.}, ``Centrality in social networks: Conceptual clarification,'' \emph{Social network: critical concepts in sociology. Londres: Routledge}, vol.~1, pp. 238--263, 2002.

\bibitem{parkes2021network}
L.~Parkes, T.~M. Moore, M.~E. Calkins, M.~Cieslak, D.~R. Roalf, D.~H. Wolf, R.~C. Gur, R.~E. Gur, T.~D. Satterthwaite, and D.~S. Bassett, ``Network controllability in transmodal cortex predicts positive psychosis spectrum symptoms,'' \emph{Biological Psychiatry}, vol.~90, no.~6, pp. 409--418, 2021.

\bibitem{kipf2016semi}
T.~N. Kipf and M.~Welling, ``Semi-supervised classification with graph convolutional networks,'' \emph{arXiv preprint arXiv:1609.02907}, 2016.

\bibitem{hamilton2017inductive}
W.~Hamilton, Z.~Ying, and J.~Leskovec, ``Inductive representation learning on large graphs,'' \emph{Advances in neural information processing systems}, vol.~30, 2017.

\bibitem{said2023enhanced}
A.~Said, M.~Shabbir, T.~Derr, W.~Abbas, and X.~Koutsoukos, ``Enhanced graph neural networks with ego-centric spectral subgraph embeddings augmentation,'' \emph{arXiv preprint arXiv:2310.12169}, 2023.

\bibitem{xu2018powerful}
K.~Xu, W.~Hu, J.~Leskovec, and S.~Jegelka, ``How powerful are graph neural networks?'' \emph{arXiv preprint arXiv:1810.00826}, 2018.

\bibitem{gilmer2017neural}
J.~Gilmer, S.~S. Schoenholz, P.~F. Riley, O.~Vinyals, and G.~E. Dahl, ``Neural message passing for quantum chemistry,'' in \emph{International conference on machine learning}.\hskip 1em plus 0.5em minus 0.4em\relax PMLR, 2017, pp. 1263--1272.

\bibitem{kriege2020survey}
N.~M. Kriege, F.~D. Johansson, and C.~Morris, ``A survey on graph kernels,'' \emph{Applied Network Science}, vol.~5, no.~1, pp. 1--42, 2020.

\bibitem{said2023survey}
A.~Said, T.~Derr, M.~Shabbir, W.~Abbas, and X.~Koutsoukos, ``A survey of graph unlearning,'' \emph{arXiv preprint arXiv:2310.02164}, 2023.

\bibitem{togninalli2019wasserstein}
M.~Togninalli, E.~Ghisu, F.~Llinares-L{\'o}pez, B.~Rieck, and K.~Borgwardt, ``Wasserstein weisfeiler-lehman graph kernels,'' \emph{Advances in neural information processing systems}, vol.~32, 2019.

\bibitem{kondor2016multiscale}
R.~Kondor and H.~Pan, ``The multiscale laplacian graph kernel,'' \emph{Advances in neural information processing systems}, vol.~29, 2016.

\bibitem{said2023augmenting}
A.~Said, M.~Shabbir, S.-U. Hassan, Z.~R. Hassan, A.~Ahmed, and X.~Koutsoukos, ``On augmenting topological graph representations for attributed graphs,'' \emph{Applied Soft Computing}, vol. 136, p. 110104, 2023.

\bibitem{bresson2017residual}
X.~Bresson and T.~Laurent, ``Residual gated graph convnets,'' \emph{arXiv preprint arXiv:1711.07553}, 2017.

\bibitem{velivckovic2017graph}
P.~Veli{\v{c}}kovi{\'c}, G.~Cucurull, A.~Casanova, A.~Romero, P.~Lio, and Y.~Bengio, ``Graph attention networks,'' \emph{arXiv preprint arXiv:1710.10903}, 2017.

\bibitem{wu2020comprehensive}
Z.~Wu, S.~Pan, F.~Chen, G.~Long, C.~Zhang, and S.~Y. Philip, ``A comprehensive survey on graph neural networks,'' \emph{IEEE transactions on neural networks and learning systems}, no.~1, 2020.

\bibitem{parkes2023using}
L.~Parkes, J.~Z. Kim, J.~Stiso, J.~K. Brynildsen, M.~Cieslak, S.~Covitz, R.~E. Gur, R.~C. Gur, F.~Pasqualetti, R.~T. Shinohara \emph{et~al.}, ``Using network control theory to study the dynamics of the structural connectome,'' \emph{bioRxiv}, 2023.

\bibitem{pasqualetti2014controllability}
F.~Pasqualetti, S.~Zampieri, and F.~Bullo, ``Controllability metrics, limitations and algorithms for complex networks,'' \emph{IEEE Transactions on Control of Network Systems}, vol.~1, no.~1, pp. 40--52, 2014.

\bibitem{ahmad2024control}
O.~U. Ahmad, A.~Said, M.~Shabbir, W.~Abbas, and X.~Koutsoukos, ``Control-based graph embeddings with data augmentation for contrastive learning,'' \emph{arXiv preprint arXiv:2403.04923}, 2024.

\bibitem{said2023neurograph}
A.~Said, R.~G. Bayrak, T.~Derr, M.~Shabbir, D.~Moyer, C.~Chang, and X.~Koutsoukos, ``Neurograph: Benchmarks for graph machine learning in brain connectomics,'' \emph{arXiv preprint arXiv:2306.06202}, 2023.

\bibitem{karateclub}
B.~Rozemberczki, O.~Kiss, and R.~Sarkar, ``{Karate Club: An API Oriented Open-source Python Framework for Unsupervised Learning on Graphs},'' in \emph{Proceedings of the 29th ACM International Conference on Information and Knowledge Management (CIKM '20)}.\hskip 1em plus 0.5em minus 0.4em\relax ACM, 2020, p. 3125–3132.

\bibitem{morris2019weisfeiler}
C.~Morris, M.~Ritzert, M.~Fey, W.~L. Hamilton, J.~E. Lenssen, G.~Rattan, and M.~Grohe, ``Weisfeiler and leman go neural: Higher-order graph neural networks,'' in \emph{Proceedings of the AAAI conference on artificial intelligence}, vol.~33, no.~01, 2019, pp. 4602--4609.

\bibitem{shi2020masked}
Y.~Shi, Z.~Huang, S.~Feng, H.~Zhong, W.~Wang, and Y.~Sun, ``Masked label prediction: Unified message passing model for semi-supervised classification,'' \emph{arXiv preprint arXiv:2009.03509}, 2020.

\bibitem{zhang2018end}
M.~Zhang, Z.~Cui, M.~Neumann, and Y.~Chen, ``An end-to-end deep learning architecture for graph classification,'' in \emph{Proceedings of the AAAI conference on artificial intelligence}, vol.~32, no.~1, 2018.

\end{thebibliography}

\end{document}